\documentclass[12pt]{article}

\PassOptionsToPackage{hidelinks,unicode,hypertexnames=false}{hyperref}

\usepackage[margin=1in]{geometry}
\usepackage[utf8]{inputenc}
\usepackage[T1]{fontenc}
\usepackage[authoryear]{natbib}
\usepackage{booktabs}
\usepackage{graphicx}
\usepackage{amsmath}
\usepackage{tabularx}
\usepackage{array}
\usepackage{longtable}
\usepackage{float}
\usepackage{nameref}
\usepackage{hyperref}
\usepackage{silence}
\newcolumntype{Y}{>{\raggedright\arraybackslash}X}

\hypersetup{
  colorlinks=true,
  linkcolor=black,
  citecolor=black,
  urlcolor=black
}

\newcommand{\seeref}[1]{\hyperref[#1]{\nameref{#1}}}

\title{From graphemic dependence to lexical structure:\\a Markovian perspective on Dante’s \emph{Commedia}}
\author{
Angelo Maria Sabatini\\
The BioRobotics Institute\\
Scuola Superiore Sant'Anna\\
Pisa, Italy\\
\texttt{angelo.sabatini@santannapisa.it}
}
\date{}

\begin{document}

\maketitle

\begin{abstract}
This study investigates the structural organisation of Dante’s \emph{Divina Commedia} through a symbolic representation based on vowel–consonant (V/C) encoding. Modelling the resulting sequence through Markov chains provides parsimonious measures of local graphemic dependence, capturing persistence and alternation patterns.

Across the poem, these measures reveal canto-level variation, with a clearer longitudinal trend from the \emph{Inferno} to the \emph{Paradiso} in the two-state compared to the four-state model. Trigram analysis identifies a restricted set of recurrent configurations acting as graphemic probes, allowing local dependency patterns to be examined in identifiable lexical environments.

A complementary classification analysis provides an independent lexical characterisation of the three cantiche and reveals a clearer longitudinal organisation, particularly within the \emph{Purgatorio}. Lexical anchoring identifies correspondences between selected graphemic probes and discriminative token features. The results provide an interpretable framework for examining correspondences between local graphemic dependence and lexical differentiation.
\end{abstract}

\section{Introduction}

Computational literary analysis combines text mining and statistical modelling with traditional close reading \citep{Piper2016}. At its core lies the idea that texts can be treated as structured sequences of symbols, amenable to quantitative description. Notably, this perspective predates the computer age and did not originally arise from attempts to interpret literary language, but from foundational work in probability theory.

The origins of this line of thought can be traced to the work of Andrej A. Markov, who, in the early twentieth century, analysed literary text as sequences of vowels and consonants to study statistical dependence between adjacent elements \citep{Markov1913}. By encoding stanzas of Aleksandr S. Puškin's \emph{Evgenij Onegin} as a binary sequence, Markov showed that probabilistic laws such as the law of large numbers could be extended to dependent sequences, thereby introducing what would later become known as Markov chains \citep{Link2006_history}. This original proposal informed later developments in information theory, most notably in the work of Claude E. Shannon \citep{Link2006_context}.

Apart from \citep{Khmelev2001} and a few other contributions \citep{Petruszewycz1983}, the direct line of influence stemming from \citep{Markov1913} did not lead to a sustained tradition in stylometry—the quantitative analysis of literary style. Stylometry has instead developed along an independent trajectory, focusing on lexical, character, syntactic, and semantic features for tasks such as authorship attribution and genre classification \citep{Stamatatos2009}. While character features, including alphabetic characters or n-grams based on aggregated frequency counts, are often employed within this framework, they are generally used as components of feature sets for classification tasks rather than as objects of sequential modelling in their own right. More recently, Markov’s ideas have been reconsidered in the context of sequential analysis and machine learning approaches \citep{Eder2016}.

Aligned with Markov’s approach, the present study adopts a deliberately minimal representation of Dante’s \emph{Divina Commedia} (hereafter \emph{Commedia}), encoding it through a vowel–consonant (V/C) representation and modelling the resulting sequence as a symbolic time series. Rather than increasing model complexity, the aim is to assess how much structural information can be recovered from local dependencies alone. 

In contrast to stylometric approaches based on aggregated counts, the analysis focuses on the longitudinal organisation of the sequence. Trigram configurations are not treated as independent features, but as locally embedded patterns whose distribution reflects the underlying dependency structure. Within this framework, the notion of \emph{memory depth} (\(MD\)) provides a statistical summary of local dependence, grounded in the use of four-state Markov chains as a generative representation of local dependency structure. A recent reappraisal of Markov’s original analysis on \emph{Evgenij Onegin} has shown that even minimal vowel–consonant encodings may provide interpretable structural signals when embedded in a contemporary statistical workflow \citep{Sabatini2026}. The approach is particularly suited to texts characterised by strong formal constraints, where metrical regularity and large-scale narrative differentiation coexist. The perspective is explicitly statistical and exploratory: the Markovian quantities derived from the V/C sequence are used to characterise local graphemic dependence and are not assumed to carry an autonomous linguistic or literary interpretation. Rather, the underlying question is how much structured variation remains detectable when a literary text is deliberately reduced to such a minimal representation.

Dante’s \emph{Commedia} has attracted a substantial body of computational work, spanning digital scholarly infrastructures, lexical and statistical analyses, metrical studies, and machine-learning–based classification. Early scholarly efforts include concordances and lexicographic resources such as \citep{Fay1888,WilkinsBergin1965}, which provided systematic access to the lexical structure of the \emph{Commedia} and anticipated later quantitative approaches to text. 

Digital projects such as the Princeton Dante Project \citep{PrincetonDanteProject}, Dante Lab \citep{DanteLab}, Digital Dante \citep{DigitalDante}, and structured digital-library initiatives aimed at semantic and encyclopaedic representation \citep{Bartalesi2017} have provided powerful platforms for accessing and annotating the text. Linguistically annotated resources include DanteSearch, a corpus of Dante’s vernacular and Latin works with lemmatisation and grammatical annotation, as well as syntactic annotation for selected works including the Commedia \citep{Tavoni2022}. More recently, the Linking Dante (LiDa) project has contributed to the interoperability of digital resources for Dante studies \citep{Concordia2024}. Other studies have explored lexical distributions \citep{Cantone2003,Lu2010}, metrical and phono-syntactic structure \citep{Asperti2021}, supervised classification approaches \citep{Khalaf2012,Saccenti2012,Romano2025}, and computational authorship verification applied to the debated attribution of the \emph{Epistle to Cangrande} \citep{Corbara2019}. While these contributions have significantly advanced the computational study of Dante, they are primarily oriented toward infrastructure, descriptive statistics, or predictive modelling, rather than toward modelling the text as a symbolic sequence in its own right and analysing its internal dependency structure.

By comparison, relatively little attention has been devoted to modelling poetic texts such as the \emph{Commedia} as sequences of elementary symbolic units, and to assessing how minimal graphemic representations capture variation in local dependency structure. Within this framework, the \emph{Commedia} is encoded as a V/C sequence and analysed as a symbolic time series to quantify local dependencies across the poem. The central contribution of the present study is not limited to the modelling framework itself, but extends to examining how the structure retained by this reduced representation relates to observable textual configurations. A simple measure of MD provides a quantitative summary of local dependency structure, while trigram patterns associated with these dependencies are examined as graphemic probes and mapped back to their lexical contexts. A classification analysis at the canto level, using cantica as the supervised label, provides an independent lexical characterisation of these predefined textual regions. Its purpose is not predictive performance per se, but the identification of lexical features that can be related to the graphemic probes through the procedure referred to here as \emph{lexical anchoring}.

These results indicate that even minimal symbolic encodings can retain systematic variation across the \emph{Commedia}. Mapping selected graphemic probes back to their lexical contexts makes it possible to examine how local dependency patterns relate to lexical differentiation among the cantiche. The resulting framework therefore provides a statistical means of exploring correspondences between different levels of textual representation without assigning an autonomous literary interpretation to the Markovian quantities themselves.

\subsection{The Divina Commedia}
Dante’s \emph{Commedia} (later referred to as the \emph{Divina Commedia} following Giovanni Boccaccio's designation) is a monumental poetic work structured into three parts (cantiche): \emph{Inferno}, \emph{Purgatorio}, and \emph{Paradiso}. Each cantica comprises 33 cantos, with an additional introductory canto at the beginning of the \emph{Inferno}, bringing the total number to 100. This tripartite organisation also makes the \emph{Commedia} a particularly informative methodological testbed. Its cantos provide repeated, structurally defined units of analysis, and membership in the three cantiche provides predefined class labels that are independent of the statistical analysis. This structure makes it possible to complement the Markovian analysis with a supervised lexical characterisation of the same textual units, and subsequently to examine correspondences between graphemic dependency patterns and lexical differentiation among the cantiche. The poem is written in hendecasyllabic verse using the \emph{terza rima} scheme, in which tercets follow an interlocking rhyme pattern (aba, bcb, cdc, ded, and so on). Each canto concludes with a single verse that continues the rhyme pattern of the final tercet. The total number of verses is 14{,}233.

The poem narrates a journey through the realms of the afterlife, from the initial disorientation in the dark wood to the final vision of God in the last canto of the \emph{Paradiso}. Dante is guided through the \emph{Inferno} and the \emph{Purgatorio} by the Latin poet Virgil, and through the \emph{Paradiso} first by Beatrice and then by St Bernard.

While the poem is unified by a consistent metrical scheme and a coherent narrative framework, the three cantiche are traditionally associated with distinct thematic and stylistic characteristics, reflecting the transition from the depiction of sin and punishment to that of purification and, ultimately, beatitude. As noted in previous studies, Dante’s style is not rigidly partitioned but rather exhibits a complex plurilingualism, rooted in the deliberate use of the vernacular Italian (\emph{volgare}) in place of Latin and in the coexistence of heterogeneous linguistic registers within a unified poetic structure \citep{Contini1970}. Nevertheless, the cantiche exhibit a differentiated yet coherent large-scale organisation, which has been discussed in relation to their stylistic and formal characteristics \citep{Auerbach1963,Ferrante1993}.

This tripartite structure makes the \emph{Commedia} a particularly suitable case for investigating large-scale variation within a single literary work. From a computational perspective, the coexistence of strong formal constraints and progressive thematic differentiation raises the question of whether such variation can be detected and quantified through minimal symbolic representations. In this sense, the poem provides an empirical testbed for exploring how local structural regularities---such as those captured by Markovian dependencies---may relate to higher-level organisation across the cantiche.

\section{Methods}

The text of the Commedia was retrieved from a publicly available GitHub repository \citep{CommediaGitHub}, which provides a structured JSON version of the poem derived from \citet{WikisourceCommedia} and based on Giorgio Petrocchi’s critical edition \emph{La Commedia secondo l’antica vulgata} (1966–67). The textual object analysed here should therefore be understood as the graphemic representation of a specified modern critical edition as transmitted through this digital source, rather than as a direct representation of an authorial orthographic state. This distinction is particularly relevant for features such as word division, apostrophes, diaeresis, and other graphical conventions that may reflect editorial choices. All analyses in this paper were conducted in R (version 4.5.2) within an exploratory observational framework. Inferential tools were used for descriptive purposes, to characterise associations and trends and identify regions of potential relevance, rather than to support confirmatory hypothesis testing. In this context, \(p\)-values and confidence intervals are interpreted as measures of compatibility with the data \citep{Wasserstein2019}. Given the exploratory setting and the limited size of the corpus, the significance level \(\alpha = 0.1\) was treated as a descriptive reference rather than as a strict decision criterion.

The JSON file was converted into a structured tibble and subjected to minimal cleaning. During preliminary corpus checking, the adopted text was compared with a separate digital transcription presented as reproducing Petrocchi’s text. This comparison was used only to identify candidate discrepancies. Where the two digital texts differed, the relevant passage was individually checked against a printed edition reproducing Petrocchi’s critical text \citep{Alighieri2017}, and patch-based replacements were applied only where this printed verification supported the correction. Particular attention was also devoted to inconsistencies in the representation of direct speech. In the digital source, direct speech was generally marked by double quotation marks, but some opening or closing delimiters were represented by single quotation marks, which could consequently be interpreted as apostrophes during the subsequent step of tokenisation (see below). Problematic instances were identified semi-automatically by searching for single quotation marks immediately preceding or following token strings. The surrounding context was then inspected to identify the corresponding direct-speech delimiters, and inconsistent instances were normalised to double quotation marks. Sample checks against the same printed edition, where direct speech is marked by guillemets, were used to verify the consistency of the normalisation. Although limited in scope, this intervention was necessary to prevent systematic misclassification of apostrophes during tokenisation. The number of corrections is negligible relative to the size of the corpus. These modifications were restricted to verified orthographic and formatting inconsistencies in the adopted digital source.

The first processing step was tokenisation. While standard tokenisers perform well in many contexts, more constrained rules are sometimes required to obtain consistent segmentations \citep{SilgeRobinson2020}. Although Italian generally uses whitespace as a word delimiter, apostrophe-bearing forms require explicit handling during lexical segmentation. Apostrophes were therefore normalised and replaced with whitespace before the final token stream was generated. This operational treatment does not depend on any classification of the apostrophe-bearing forms. The tokeniser released with the analysis pipeline also records, for each apostrophe, a rule-based operational label describing its local context; these labels are not used in the present analyses and are not intended as a linguistic taxonomy.

The tokeniser enabled the construction of a Dante-specific stopword list. Standard Italian stopword lists, designed for modern prose, do not adequately capture the variability of \emph{volgare}, particularly in the presence of apostrophised function words (e.g. \emph{l', ch', s'}). Candidate stopwords were identified based on frequency and dispersion across cantos. Preference was given to forms that are both frequent and widely distributed, including some highly frequent pronominal and adjectival forms. The list was then manually refined to retain grammatical function words while excluding content-bearing terms. The final stopword list is provided in Supplementary Material~A. Very short tokens (one or two characters) that may arise from the tokenisation process were handled in subsequent filtering steps when constructing the vocabulary for classification tasks.

Following tokenisation, the text was analysed at the character level. For each canto, punctuation and non-alphabetic symbols were removed, and sequences of alphabetic characters were extracted to enable consistent character-level comparisons across the poem. Characters marked with diaeresis were retained. In addition, two simple descriptors were computed at the canto level: mean token length, expressed as the number of characters per token, and mean number of tokens per verse. The latter was obtained by dividing the number of tokens in each canto by its number of verses. The two measures provide complementary descriptions of the lexical organisation of the text: mean token length is directly sensitive to the character-level representation of individual tokens; the mean number of tokens per verse describes their distribution within the common metrical frame of the hendecasyllabic verse. Unless otherwise noted, correlations are computed using Spearman’s rank coefficient $\rho$.

The text was then encoded as a binary sequence of vowels (V) and consonants (C), mapping each character to one of these two categories. This representation abstracts from lexical content while preserving local patterns, enabling the analysis of short-range structure independently of semantic content. The encoding is strictly graphemic rather than phonetic: characters marked with diaeresis retain their vowel-class membership and are therefore mapped to V in the same way as the corresponding unmarked vowels. The mapping follows written character classes rather than phonological function; thus, for example, \emph{u} in \emph{qu} and \emph{i} in sequences such as \emph{gli} or \emph{cia} are encoded as vowels irrespective of their phonetic realisation.

\begin{figure}[htbp]
\centering
\includegraphics[width=0.9\linewidth]{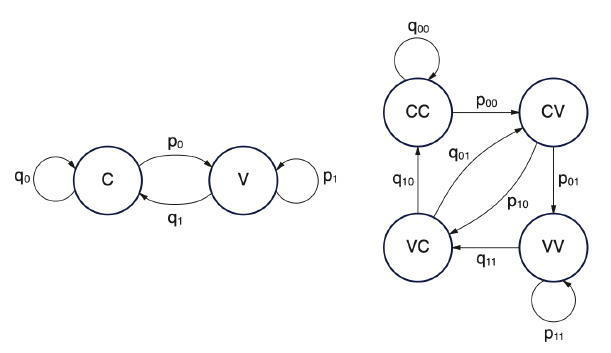}
\caption{\textbf{Two- and four-state Markov representations of the vowel--consonant (V/C) sequence}. Left: two-state model (V, C) with first-order transition probabilities. Right: four-state model based on overlapping symbol pairs (VV, VC, CV, CC), capturing dependencies at the trigram level. The transition probabilities (\(p_{ij}\), \(q_{ij}\)) define the local structure of the sequence.}
\label{fig:fig-fig1}
\end{figure}

Two-state and four-state Markov chain models were considered (Fig.~\ref{fig:fig-fig1}). The four-state representation, defined on overlapping symbol pairs (VV, VC, CV, CC), captures dependencies across three consecutive characters and provides a generative description of local dependency structure, preserving transition dynamics between adjacent symbol pairs \citep{Sabatini2026}.
Transition probabilities were estimated separately for each canto, which served as the primary statistical unit of analysis, providing a structurally defined and sufficiently large segment for stable estimation of local dependency patterns.

Dependence in the sequence was quantified using the dispersion coefficient \(CF\) introduced in Markov's analysis \citep{Markov1913}. This coefficient measures the deviation of the observed variability of a binary sequence from that expected under independence: \(CF = 1\) under independence, whereas values below or above \(1\) indicate reduced or inflated variability, respectively. In modern statistical terms, it can be interpreted as a measure of serial dependence, analogous to a variance inflation factor in the presence of autocorrelation. For example, in languages such as Italian, where vowels and consonants tend to alternate, a vowel is more likely to follow a consonant (and vice versa) than expected under independence; as a result, the occurrence of vowels is more regular than independence would predict, and \(CF < 1\) \citep{Markov1913, Sabatini2026}. In this study, \(MD\) is defined as \(1 - CF\); since \(CF < 1\) in all cantos, higher values of \(MD\) correspond to stronger deviations from independence in the V/C sequence.

In addition to the four-state representation, a two-state Markov model was used to capture global dependence in the sequence. The corresponding dispersion coefficient (hereafter \(CF_{\text{simple}}\)) provides a coarse-grained measure of serial dependence, whereas the four-state coefficient \(CF\) captures finer local structure through transitions between symbol pairs.

Although derived from different state spaces, \(CF_{\text{simple}}\) and \(CF\) (and the corresponding \(MD\) values) capture related aspects of departure from independence and can be interpreted as complementary projections of the same dependency structure. The two-state measure is primarily driven by first-order transition probabilities (\(p_1, p_0\)), whereas the four-state coefficient also incorporates second-order effects expressed through trigram-equivalent transitions (\(p_{11}, p_{10}, p_{01}, q_{00}\)). These transition probabilities admit a direct interpretation in terms of trigram structure: self-persistence terms (\(p_{11}, q_{00}\)) correspond to configurations without transitions, whereas transition terms such as \(p_{10}, p_{01}\) capture configurations involving one or more transitions. The position of these transitions within the trigram is made explicit in the classification introduced below.

A sensitivity analysis was conducted to examine the relationship between \(MD\) and the transition probabilities of the underlying four-state Markov representation. Following the framework introduced in \citep{Sabatini2026}, particular attention was given to probabilities describing self-persistence (\(p_{11}\) and \(q_{00}\)), together with the corresponding complementary transitions. At the same time, first-order probabilities (\(p_0\), \(p_1\)) and transition terms (\(p_{10}\), \(p_{01}\), and the corresponding complementary transitions) were also considered as part of the general transition structure.

Two levels of trigram analysis were distinguished. All letter trigrams occurring in the text were extracted and each was classified according to its vowel–consonant composition into one of eight V/C configurations (VVV, VVC, CCC, CCV, VCC, VCV, CVV, CVC); for example, \emph{nte} and \emph{sto} are both instances of CCV. This set exhausts the space of possible trigram patterns in the binary encoding and corresponds to the second-order transition structure of the four-state Markov representation. Configurations were further classified according to the number and position of transitions: no transition (CCC, VVV), a single transition at the end (CCV, VVC), a single transition at the start (CVV, VCC), and two transitions (CVC, VCV). For convenience, these classes are denoted as 0, 1E, 1S, and 2, respectively. Each occurrence was mapped back to its position in the adopted textual representation, allowing the surrounding text and its lexical realisation to be retrieved.

Occurrences were aggregated at the canto level, both for each letter trigram and, by summation, for each V/C configuration. Frequency profiles across cantos were assessed using Spearman rank correlation to evaluate the presence of systematic monotonic associations with canto position. This choice was guided by exploratory inspection of the data, which suggested broadly monotonic relationships without supporting more restrictive parametric assumptions.

Letter trigrams were then selected as \emph{graphemic probes}: observable instances of the second-order dependencies in the V/C sequence, which can be traced back to the words and word boundaries in which they occur. Candidate probes were retained when their frequency was monotonically associated with canto position (Spearman's \(\rho\), \(p < 0.1\)) and when the direction of this association was consistent with the trend observed for the corresponding trigram class at the level of V/C configurations. Discordant cases were set aside.

\emph{Lexical anchoring} refers to the extent to which a given probe admits stable realisations within individual lexical units, as opposed to being primarily expressed across word boundaries. This distinction is not absolute: some configurations occur both within and across lexical units, while others are inherently associated with cross-word realisations. In practice, it is adopted as a simplifying operational framework to support subsequent analysis based on single-token representations. While this choice may discard part of the boundary-level signal, it retains sufficient structure to connect probes to their instantiation in the text.

A supervised classification analysis was conducted at the canto level, using the three cantiche as reference classes. The aim of this step is not to maximise predictive performance, but to identify terms that provide an interpretable characterisation of the cantiche. Accordingly, classification is used as an exploratory and descriptive tool rather than as a machine learning task in the conventional sense, supporting the operationalisation of lexical anchoring by identifying relevant features associated with the graphemic probes.

Two classifiers were employed in parallel: sparse partial least squares discriminant analysis (sPLS-DA), implemented using the \texttt{mixOmics} package \citep{Rohart2017}, and multinomial logistic regression with elastic-net regularisation (EN-MNLR), fitted within the \texttt{tidymodels} framework via \texttt{glmnet} \citep{KuhnSilge2022}. These models were selected for their complementary interpretive properties. sPLS-DA provides a low-dimensional projection that facilitates the visualisation of global lexical structure, while EN-MNLR yields direct coefficient-based rankings of class-specific features. Rather than producing identical feature rankings, the two models identify partially overlapping sets of informative terms. Their combined use therefore broadens the coverage of lexical signals associated with each cantica, while maintaining an explicitly interpretive, rather than performance-driven, perspective.

The feature space was constructed from the tokenised text produced by the custom tokeniser described above. After lowercasing, only alphabetic tokens were retained, and the Dante-specific stopword list was applied. For each canto, a document--term matrix was constructed using term-frequency counts. Features were defined from observed token forms rather than lemmatised lexical units; consequently, graphically distinct forms were retained as distinct features. To maintain a compact and interpretable representation, the vocabulary was restricted to the most frequent tokens in the corpus (top-500), in line with common practice in stylometric analyses \citep{Saccenti2012, HaCohenKerner2020}. A minimum token-length threshold ($\geq 3$ characters) was also applied during feature construction, reducing noise from very short forms while preserving short but stylistically informative terms (e.g., \emph{ahi, piè, ivi}) that may contribute to the analysis. 

For both classifiers, model selection was based on Monte Carlo validation using the same class-stratified train/test splits (80/20). To avoid information leakage, all preprocessing steps, including vocabulary selection and parameter tuning, were performed within each training fold \citep{KuhnSilge2022}. In each iteration, the number of retained variables per component (\texttt{keepX}) for sPLS-DA and the penalty (\(\lambda\)) and mixing (\(\alpha\)) parameters for EN-MNLR were tuned by internal cross-validation. Model performance was summarised across runs using accuracy (Acc.), balanced accuracy (Bal. Acc.), macro-averaged F1 score (F1), and Matthews correlation coefficient (MCC) \citep{Romano2025}. For sPLS-DA, the number of components was fixed to two, consistent with the three-class structure of the target variable, and top terms were interpreted based on the magnitude of their loadings in the discriminant space. As sPLS-DA does not yield class-specific coefficients, terms were assigned to cantiche based on the sign of their loading values. In the case of EN-MNLR, class-specific coefficients were used to identify the most informative terms for each cantica.

The resulting feature rankings were combined into a unified representation, highlighting both concordant and model-specific signals. Terms identified by both models were reported without annotation, whereas those selected exclusively by one model were marked accordingly. Lexical anchoring was then defined operationally as a mapping between two independently derived representations: the graphemic probes selected from the analysis of V/C configurations and the token features identified by the supervised models as discriminative for the cantiche. For each high-ranking token feature, the occurrence of a selected probe as an intra-token realisation was identified, and its frequency and relative concentration across cantiche were combined into a simple signal score used to rank the resulting probe–token associations. The procedure therefore restricts the search for lexical realisations of the selected probes to a model-derived set of discriminative token features, rather than treating the occurrence of a probe in any lexical item as evidence of lexical specificity.

\section{Results}
The results are organised to progressively connect descriptive properties of the text with the analysis of local dependency patterns and their relation to lexical organisation. The initial part summarises a small set of descriptive and structural features that provide context for the symbolic encoding. The subsequent sections examine how these patterns, captured by the Markov representation, vary across the poem and give rise to graphemic probes, which are then linked to lexical anchors and higher-level textual organisation.

Table~\ref{tab:tab-tab2} provides a representative example of the tokenisation procedure applied to the text, including the operational handling of apostrophe-bearing forms during segmentation. Apostrophe-bearing forms are included for descriptive purposes, but the core analysis relies on the symbolic sequence produced by segmentation.

\begin{table}[htbp]
\centering
\caption{\textbf{Tokeniser output for tercet 4, canto I, \emph{Inferno}}. Lexical tokens are reported (punctuation and quotation marks excluded); apostrophe-bearing forms are segmented according to the operational tokenisation rules described in the Methods section.}
\label{tab:tab-tab2}
\vspace{0.2em}
\begin{tabular}{@{}l l@{}}
\toprule
\textbf{Original form} & \textbf{Token} \\
\midrule
\textit{Io} & \textit{Io} \\
\textit{non} & \textit{non} \\
\textit{so} & \textit{so} \\
\textit{ben} & \textit{ben} \\
\textit{ridir} & \textit{ridir} \\

\textit{com’i’} & \textit{com}  \\
                & \textit{i}  \\

\textit{v’intrai} & \textit{v}  \\
                 & \textit{intrai}  \\

\midrule

\textit{tant’era} & \textit{tant} \\
                  & \textit{era}  \\

\textit{pien} & \textit{pien} \\
\textit{di} & \textit{di} \\
\textit{sonno} & \textit{sonno} \\
\textit{a} & \textit{a}  \\
\textit{quel} & \textit{quel} \\
\textit{punto} & \textit{punto} \\

\midrule

\textit{che} & \textit{che} \\
\textit{la} & \textit{la} \\
\textit{verace} & \textit{verace} \\
\textit{via} & \textit{via} \\
\textit{abbandonai} & \textit{abbandonai} \\

\bottomrule
\end{tabular}
\end{table}

Mean token length shows a slight increase across cantos (\(\rho=0.39, p<0.001\)); the mean number of tokens per verse shows a corresponding decrease (\(\rho=-0.48, p<0.001\)). The two descriptors are strongly negatively associated (\(\rho=-0.94, p<0.001\)). Mean token length increases from the \emph{Inferno} (\(3.93\) characters per token) through the \emph{Purgatorio} (\(3.96\)) to the \emph{Paradiso} (\(4.01\)), while the mean number of tokens per verse decreases from \(7.23\) to \(7.16\) and \(7.02\), respectively. Differences across cantiche are observed for both descriptors (Kruskal–Wallis test: \(p<0.001\) for mean token length; \(p<0.00001\) for mean number of tokens per verse).

\begin{figure}[htbp]
\centering
\includegraphics[width=0.8\linewidth]{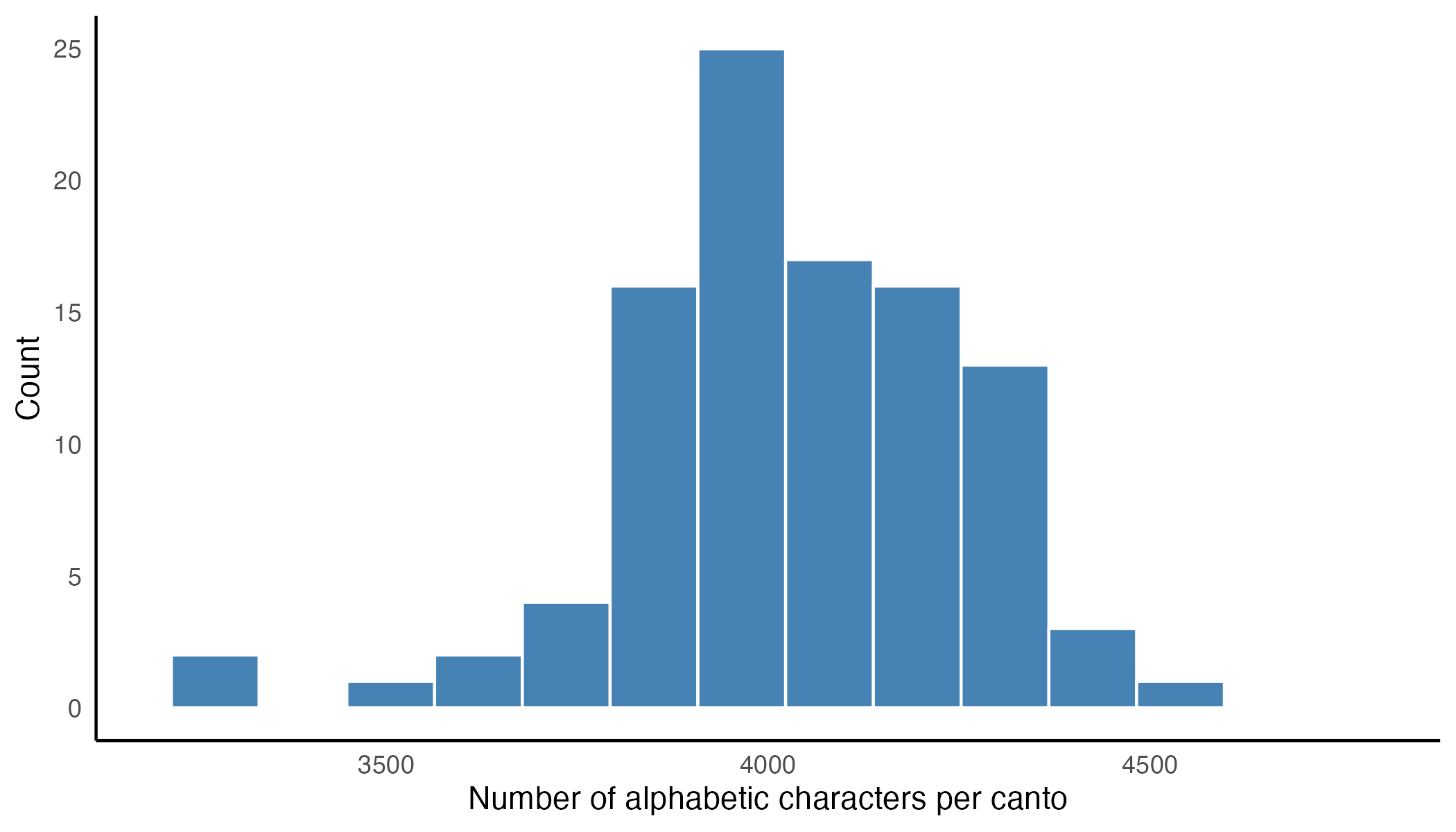}
\caption{\textbf{Alphabetic character counts per canto}. Distribution of alphabetic characters after removing punctuation and non-alphabetic symbols.}
\label{fig:fig-fig3}
\end{figure}

Figure~\ref{fig:fig-fig3} summarises the number of alphabetic characters per canto after removing punctuation and non-alphabetic symbols. Cantos contain on average about 4{,}000 characters (mean \(= 4{,}028\), SD \(= 219\)), with limited variability (coefficient of variation \(\approx 5.5\%\)). This relatively consistent and sufficiently large length provides a favourable basis for estimating transition probabilities at the canto level \citep{Sabatini2026}.
Transition probabilities were estimated separately for each canto, which served as the primary unit of analysis.

\begin{figure}[htbp]
\centering
\includegraphics[width=0.8\linewidth]{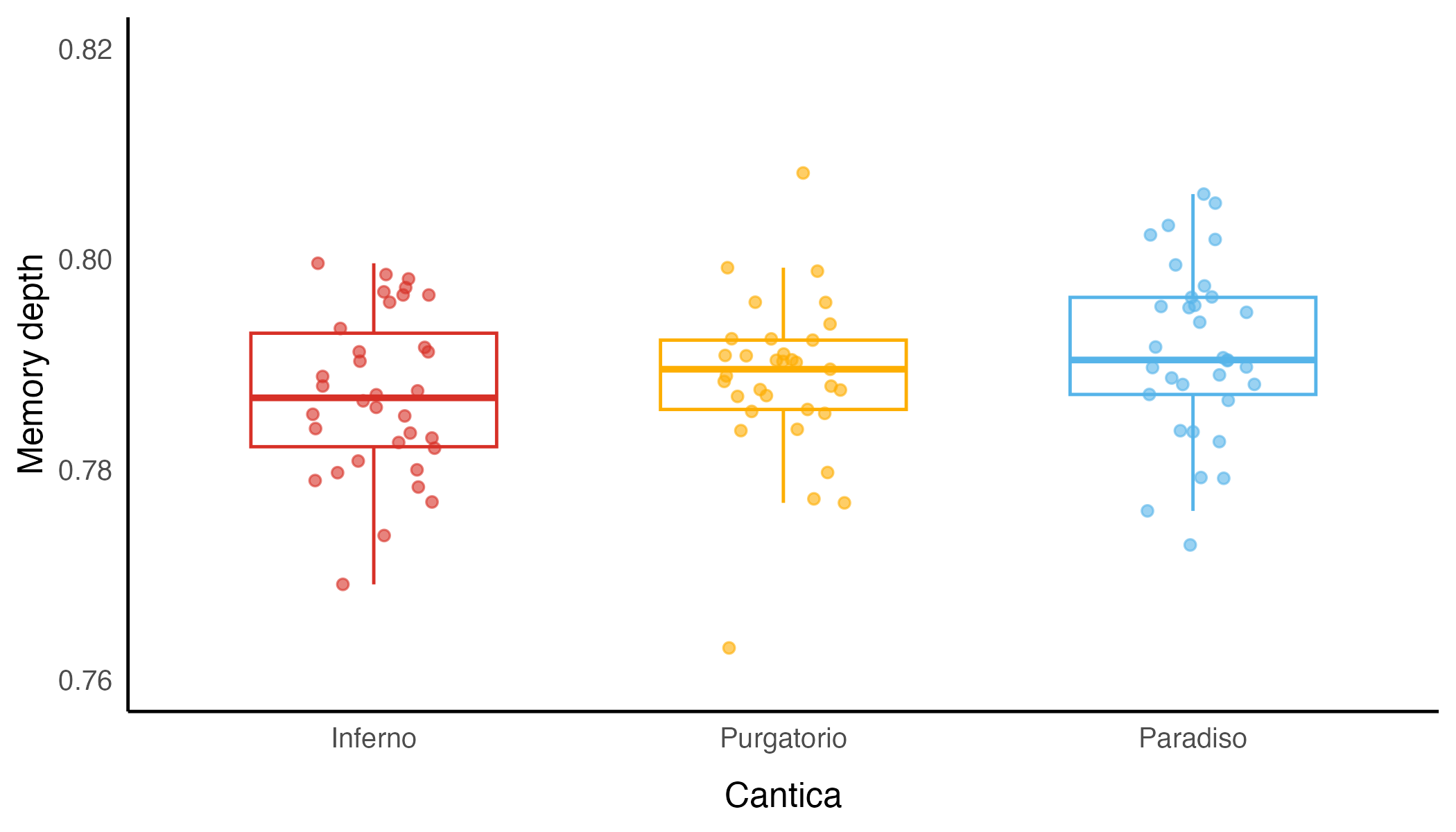}
\caption{\textbf{Memory depth across cantiche.} Boxplots show the distribution of \(MD\) values across cantos, estimated using the four-state Markov model; points represent individual cantos.}
\label{fig:fig-fig4}
\end{figure}

Figure~\ref{fig:fig-fig4} summarises the distribution of \(MD\) values estimated using the four-state Markov model. Values are concentrated in a narrow range (0.76--0.81), indicating that the local dependency structure is broadly stable across the poem. At the same time, a slight upward shift in the distribution from the \emph{Inferno} to the \emph{Paradiso} indicates a weak but systematic large-scale modulation.
Although the dispersion of \(MD\) values is limited, both the two-state and four-state representations reveal interpretable differences across the poem. Differences across cantiche were assessed using the Kruskal--Wallis test, followed by Dunn's post-hoc comparisons with Holm correction. 

\(MD\) estimated from the two-state Markov model shows a clear increasing trend across cantos (linear slope \(= 1.63 \times 10^{-4}\) per canto, \(p = 7.6 \times 10^{-5}\); \(\rho = 0.39\), \(p = 5.0 \times 10^{-5}\)). Differences across cantiche are statistically significant (Kruskal--Wallis test: \(p = 4.1 \times 10^{-5}\)). Post-hoc comparisons indicate higher values in the \emph{Paradiso} relative to both the \emph{Inferno} (\(p < 0.001\)) and the \emph{Purgatorio} (\(p < 0.01\)), while no significant difference is observed between the \emph{Inferno} and the \emph{Purgatorio} (\(p = 0.155\)). This pattern suggests that the increase in dependence is driven primarily by higher values in the \emph{Paradiso}, rather than by a smooth progression across all three cantiche. The same qualitative pattern is observed in the four-state representation, but with reduced effect size and weaker statistical support. A positive trend is detected (slope \(= 6.6 \times 10^{-5}\) per canto, \(p = 0.017\)), accompanied by a weaker monotonic association (\(\rho = 0.23\), \(p = 0.024\)). However, differences across cantiche do not reach statistical significance (Kruskal--Wallis test: \(p = 0.131\)).

The association between \(MD\) and the transition probabilities involved in its computation was examined. Correlations with \(MD\) are reported as partial Spearman coefficients. Negative associations were observed between \(MD\) and the persistence-related transitions \(p_{11}\) (partial \(\rho = -0.63\), \(p < 10^{-11}\)) and \(q_{00}\) (partial \(\rho = -0.68\), \(p < 10^{-13}\)). A weaker negative association was found for \(p_1\) (partial \(\rho = -0.35\), \(p < 0.001\)), while no clear monotonic relationships were observed for \(p_0\), \(p_{10}\), and \(p_{01}\).

Differences across cantiche were assessed for each transition component using the same procedure described above. No statistically significant differences were observed for the self-persistence terms \(p_{11}\) and \(q_{00}\) (Kruskal--Wallis tests: \(p = 0.45\) and \(p = 0.59\), respectively), indicating limited variation across cantiche. In contrast, the probability \(p_{10}\) varies across cantiche (Kruskal--Wallis test: \(p < 10^{-4}\)), with higher values in the \emph{Paradiso} compared to the \emph{Inferno} (\(p < 0.001\)) and the \emph{Purgatorio} (\(p < 0.05\)). Similarly, \(p_{01}\) varies across cantiche (Kruskal--Wallis test: \(p < 0.005\)), showing a significant increase from the \emph{Purgatorio} to the \emph{Paradiso} (\(p < 0.01\)). This contrast between relatively stable persistence terms and varying transition probabilities suggests that changes in \(MD\) are not driven by a single transition component. Finally, the probabilities \(p_0\) and \(p_1\) also vary across cantiche (Kruskal--Wallis tests: \(p < 10^{-4}\) and \(p = 0.009\), respectively), with higher values observed in the \emph{Paradiso}.

Trigram configurations were screened for frequency profiles showing monotonic variation across cantos and interpreted according to the classification defined in the Methods section. This classification links trigram behaviour to the variation patterns of the corresponding transition probabilities and, ultimately, to the behaviour of \(MD\). For ease of reference, trigram classes are denoted as 0, 1E, 1S, and 2.

A graphemic probe was identified when trigram counts exhibited an association with canto position (\(p < 0.1\)) and behaviour was consistent with the expected direction of the corresponding trigram class. Trigrams showing divergent trends were set aside and not considered further. Occurrences of the selected probes were then classified according to whether they appear within a single lexical unit or span across word boundaries. At the aggregate level, configurations 0, 1E and 1S are predominantly realised within single words
(77.0\%, 80.0\%, and 72.1\%, respectively), indicating a strong association with intra-lexical structure. In contrast, configuration 2 is more frequently realised across word boundaries (57.4\%), suggesting a stronger link with boundary-level interactions between adjacent tokens.
\begin{table}[htbp]
\centering
\caption{\textbf{Selected graphemic probes and their distribution across lexical contexts}. For each probe, the corresponding trigram type, class (0 = no transition; 1E = one transition at the end; 1S = one transition at the start; 2 = two transitions), and the proportion of occurrences within single lexical units (SW\%) are reported.}
\label{tab:tab-tab3}
\begin{tabular}{@{}l c c c@{}}
\toprule
Probe & Type & Class & SW (\%) \\
\midrule
str & CCC & 0  & 99.9 \\
sto & CCV & 1E & 99.4 \\
nto & CCV & 1E & 97.1 \\
uel & VVC & 1E & 95.7 \\
com & CVC & 2  & 95.5 \\
che & CCV & 1E & 95.2 \\
cia & CVV & 1S & 94.9 \\
ues & VVC & 1E & 92.3 \\
lla & CCV & 1E & 92.1 \\
and & VCC & 1S & 92.0 \\
nte & CCV & 1E & 90.5 \\
est & VCC & 1S & 88.1 \\
tan & CVC & 2  & 85.4 \\
ome & VCV & 2  & 84.3 \\
nde & CCV & 1E & 75.7 \\
noi & CVV & 1S & 62.4 \\
ion & VVC & 1E & 62.4 \\
ede & VCV & 2  & 55.3 \\
nel & CVC & 2  & 53.4 \\
del & CVC & 2  & 43.1 \\
eco & VCV & 2  & 15.1 \\
ich & VCC & 1S & 8.0 \\
ioc & VVC & 1E & 6.6 \\
ela & VCV & 2  & 4.2 \\
equ & VCV & 2  & 1.6 \\
hel & CVC & 2  & 1.5 \\
ain & VVC & 1E & 1.0 \\
ein & VVC & 1E & 0.2 \\
eio & VVV & 0  & 0.0 \\
\bottomrule
\end{tabular}
\end{table}
Some probes are primarily realised within lexical units (e.g., \emph{mae\textbf{str}o, que\textbf{sto}, giu\textbf{nto}}), while others arise across word boundaries, as in sequences such as \emph{rovina v\textbf{a in} basso} or \emph{sempr\textbf{e in} quell'aura}. The context \emph{di pietad\textbf{e io} venni} illustrates a probe that occurs exclusively across word boundaries and does not admit intra-lexical realisations (SW\% = 0). This distinction reflects two complementary contexts in which the observed trigram structures occur: intra-word configurations and boundary-level interactions between adjacent words. Consistent with the aggregate analysis, configurations involving multiple transitions are more frequently realised across word boundaries, whereas those with zero or a single transition tend to be anchored within lexical units. Probes dominated by cross-word occurrences are therefore less likely to be retained under lexical anchoring procedures, which privilege stable intra-word configurations.

\begin{figure}[htbp]
\centering
\begin{minipage}{0.48\textwidth}	
    \centering
    \includegraphics[width=\linewidth]{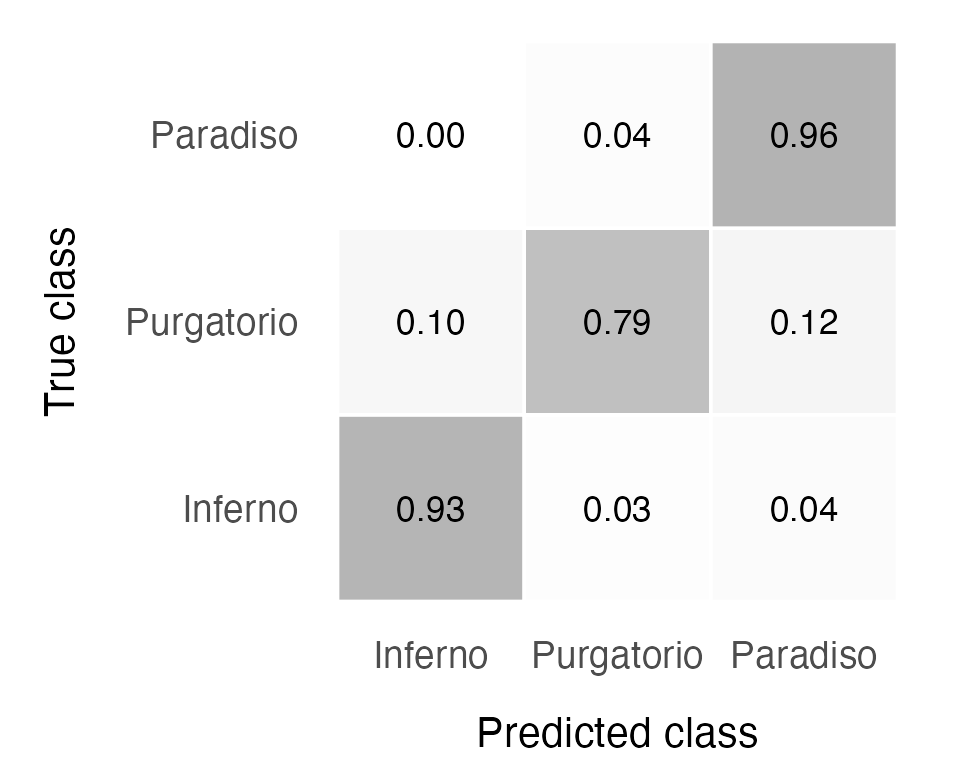}
    {\small (a) sPLS-DA}
    \vspace{0.3em}
\end{minipage}
\hfill
\begin{minipage}{0.48\textwidth}
    \centering
    \includegraphics[width=\linewidth]{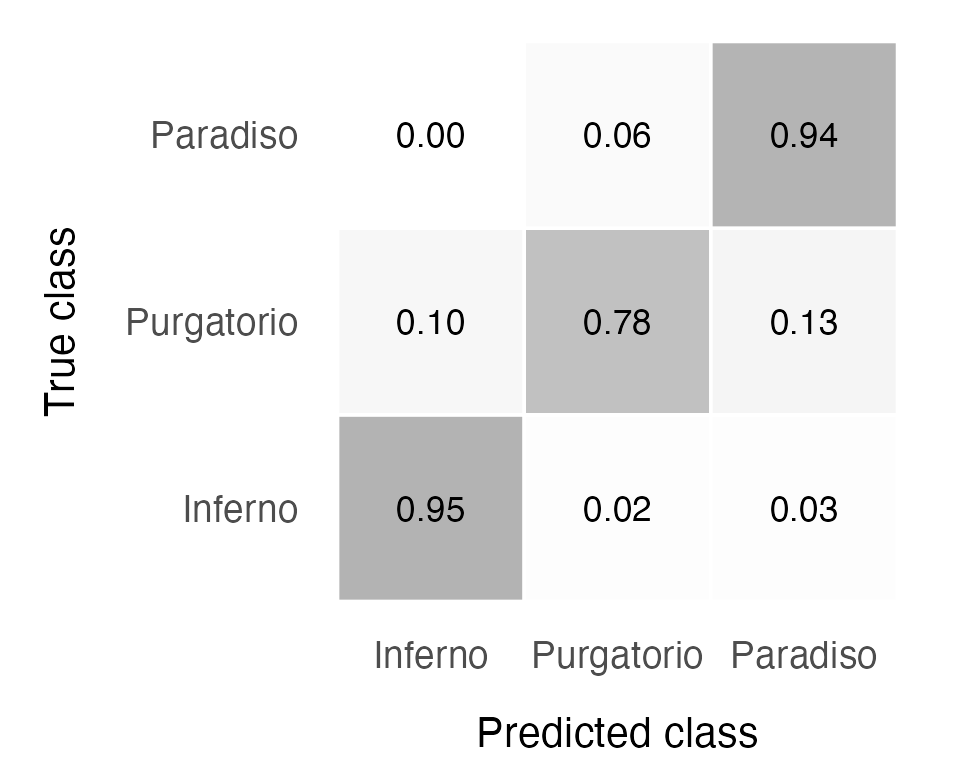}   
    {\small (b) EN-MNLR}
    \vspace{0.3em}
\end{minipage}
\caption{\textbf{Out-of-sample confusion matrices for cantica-level classification}. Values are normalised by true class, allowing direct comparison of class-wise performance.}
\label{fig:fig-fig5}
\end{figure}

\begin{figure}[htbp]
\centering
\includegraphics[width=0.75\linewidth]{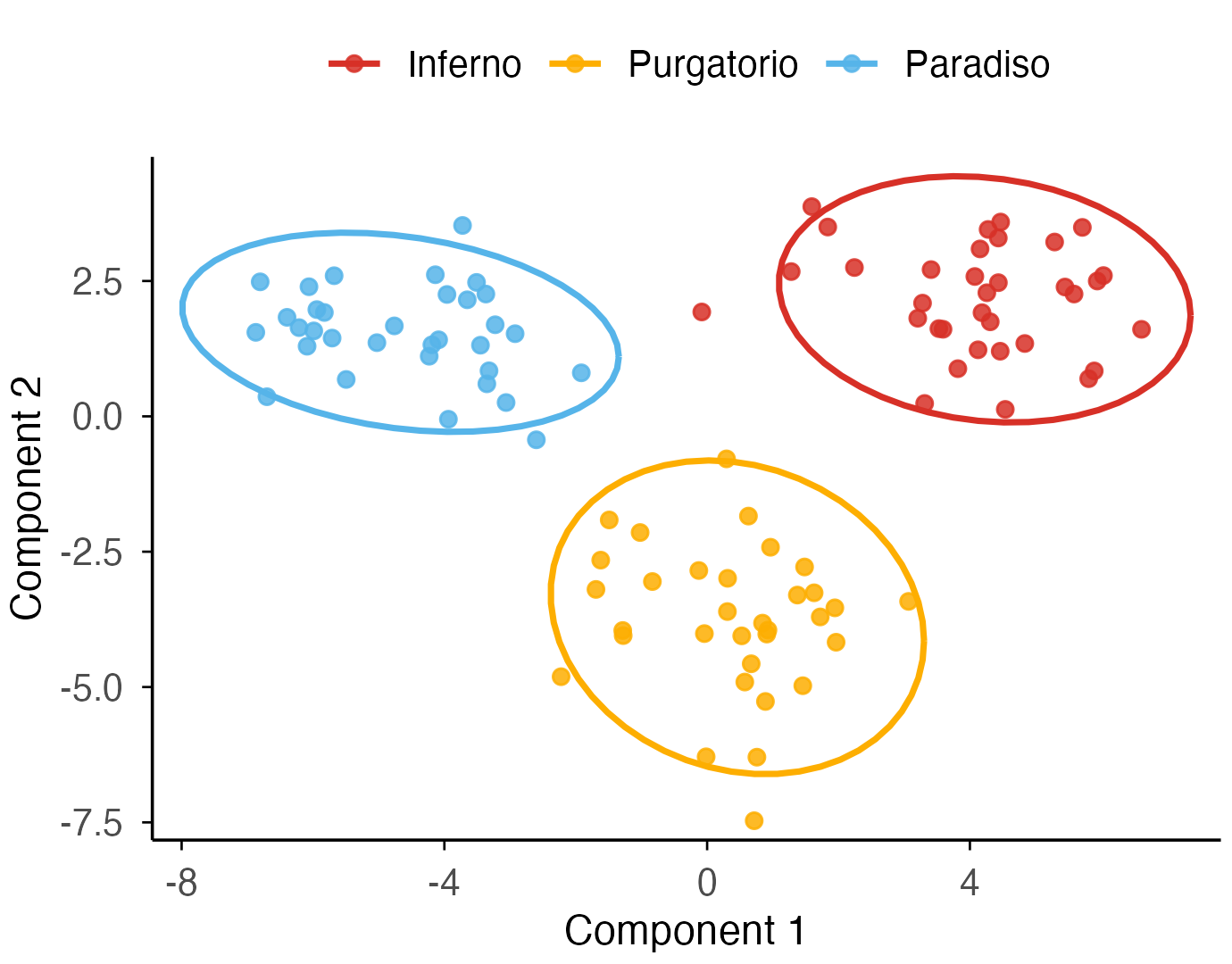}
\caption{\textbf{sPLS-DA latent space for canto-level lexical profiles}. The two discriminant components separate the three cantiche in the lexical feature space. Points represent individual cantos; ellipses indicate 95\% concentration regions. The projection is obtained from the sPLS-DA model selected during tuning and fitted on the full dataset.}
\label{fig:fig-fig6}
\end{figure}

\begin{figure}[htbp]
\centering
\begin{minipage}{0.48\textwidth}
    \centering
    \includegraphics[width=0.95\linewidth]{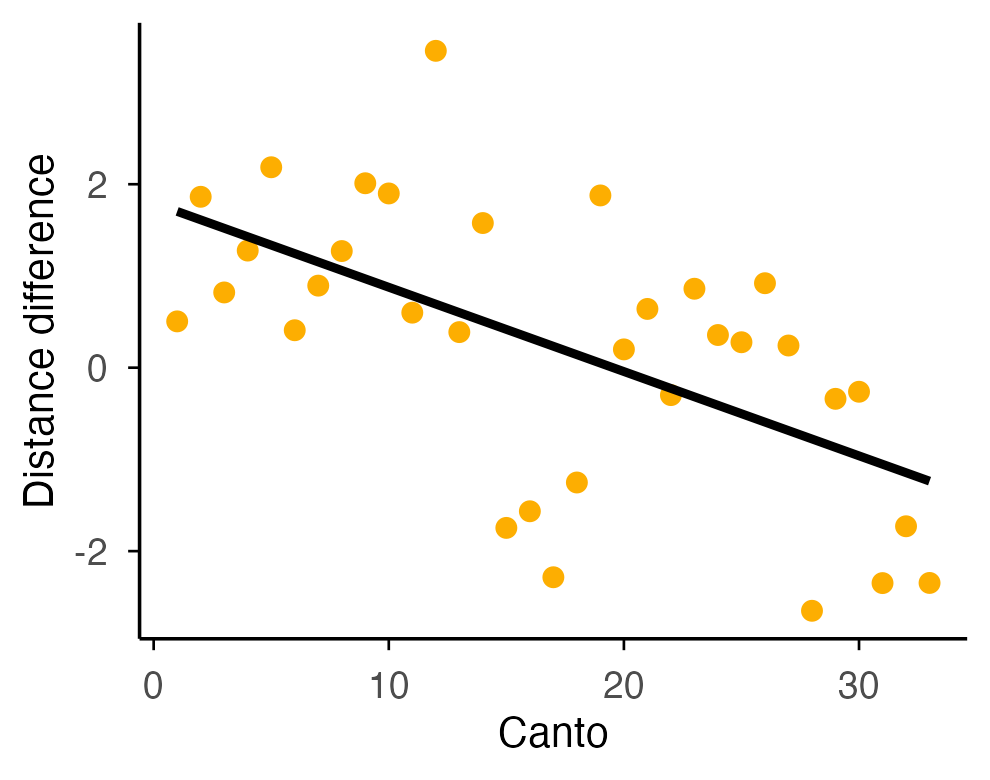}
    {\small (a) sPLS-DA}
     \vspace{0.2em}
\end{minipage}
\hfill
\begin{minipage}{0.48\textwidth}
    \centering
    \includegraphics[width=0.95\linewidth]{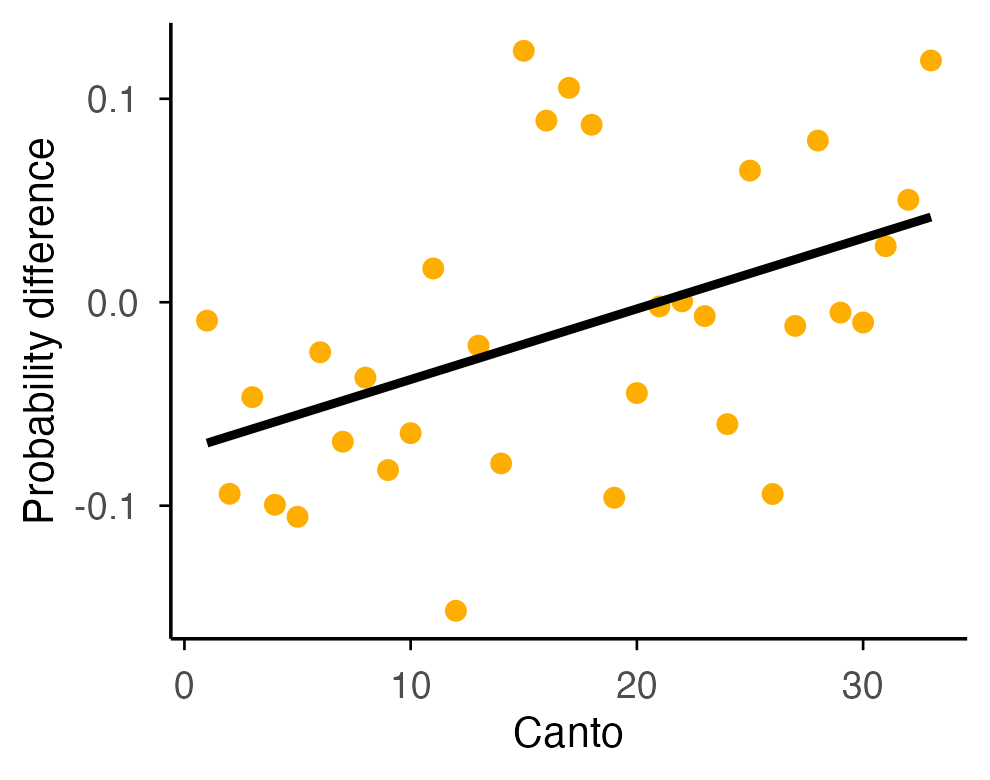}
   {\small (b) EN-MNLR}
    \vspace{0.2em}
\end{minipage}
\caption{\textbf{Internal progression of the \emph{Purgatorio}}. (a) sPLS-DA: signed distance difference. (b) EN-MNLR: signed probability difference. Differences are computed as \emph{Paradiso} minus \emph{Inferno}. The two representations exhibit consistent monotonic trends, with opposite signs reflecting their respective definitions.}
\label{fig:fig-fig7}
\end{figure}

Supervised classification models were trained using the cantica as target variable. Classification performance for both models is reported in Table~\ref{tab:tab-tab4}. The optimal sPLS-DA model consistently selected approximately 146--210 variables per component, while the EN-MNLR model converged to a near-ridge solution at the boundary of the tuning grid (\(\lambda = 10^{-6}, \alpha = 0\)). This behaviour indicates that, within the adopted tuning grid, only minimal regularisation was selected by the EN-MNLR model.
\begin{table}[htbp] 
\centering 
\caption{\textbf{Classification performance of sPLS-DA and EN-MNLR models.} Metrics are reported as mean $\pm$ standard deviation across Monte Carlo validation runs (100 iterations; 80/20 stratified splits).}
\label{tab:tab-tab4}
\vspace{0.1em}
\begin{tabular}{@{}l c c c c@{}}
\toprule
\textbf{Classifier} & \textbf{Acc.} & \textbf{F1} & \textbf{MCC} & \textbf{Bal. Acc.} \\ 
\midrule 
sPLS-DA 
& $0.892 \pm 0.064$ 
& $0.890 \pm 0.065$ 
& $0.847 \pm 0.091$ 
& $0.919 \pm 0.048$ \\ 

EN-MNLR 
& $0.888 \pm 0.063$ 
& $0.885 \pm 0.065$ 
& $0.839 \pm 0.092$ 
& $0.916 \pm 0.047$ \\ 
\bottomrule 
\end{tabular} 
\end{table}
Figure~\ref{fig:fig-fig5} shows the out-of-sample confusion matrices for the two classifiers. Values are normalised by true class, allowing direct comparison of class-wise performance. Both models exhibit clear separation of the three cantiche. Predictions for the \emph{Inferno} and the \emph{Paradiso} are almost entirely accurate, with negligible confusion between the two, while misclassification is concentrated in the \emph{Purgatorio}, which occupies an intermediate position and is occasionally assigned to either neighbouring cantica. The two models display highly consistent patterns, with only minor differences in the distribution of errors. 
Figure~\ref{fig:fig-fig6} shows the representation defined by the two discriminant components of the sPLS-DA model, where the three cantiche form distinct regions in the latent space.
Figure~\ref{fig:fig-fig7} examines the internal progression of the \emph{Purgatorio} across cantos. For each canto, a signed difference is computed between its relative proximity to  the \emph{Paradiso} and the \emph{Inferno}. This difference is evaluated using two independent representations: centroid distances in the sPLS-DA space and predicted class probabilities from the EN-MNLR model. Both representations reveal a consistent monotonic trend across cantos, with opposite signs reflecting their definition. In the sPLS-DA space, the distance difference decreases significantly along the sequence (slope \(= -0.092\) per canto, \(p < 0.0005\); \(\rho = -0.64\), \(p < 0.0001\)), indicating a progressive shift towards the \emph{Paradiso}. Consistently, the EN-MNLR probabilities show a significant increase in the same direction (slope \(= 0.003\) per canto, \(p < 0.01\); \(\rho = 0.50\), \(p < 0.005\)).
Both representations reveal a clear and statistically significant monotonic trend. In the sPLS-DA space, the distance difference decreases steadily across cantos, indicating a progressive shift towards the \emph{Paradiso}, while the EN-MNLR probabilities show a corresponding increase. These results indicate that the \emph{Purgatorio} exhibits a structured internal progression, rather than behaving as a homogeneous intermediate category. The agreement between the two modelling frameworks indicates that this pattern is detectable under both lexical representations.

Table~\ref{tab:tab-tab5} reports the highest-ranking terms obtained from the EN-MNLR model, grouped by cantica. Within this deliberately interpretable bag-of-words setting, these terms provide a direct lexical counterpart to the previously identified probes, linking local graphemic configurations to discriminative lexical features across the cantiche. Terms identified by both classifiers are marked accordingly, while additional contributions from sPLS-DA are indicated separately. Notably, \emph{Beatrice} and the editorially marked form \emph{Bëatrice} are retained as distinct observed token features and both appear among the top-ranked terms.

\renewcommand{\arraystretch}{1.1}
\begin{table}[htbp]
\centering
\caption{\textbf{Top-ranked lexical terms by cantica}. Terms are ranked according to the EN-MNLR model. Terms marked with \(^{\dagger}\) are also identified by sPLS-DA, whereas those marked with \(^{\ddagger}\) are identified exclusively by sPLS-DA.}
\label{tab:tab-tab5}
%\footnotesize
\begin{tabular}{@{}l @{\hspace{10pt}} l @{\hspace{10pt}} l@{}}
\toprule
\emph{Inferno} & \emph{Purgatorio} & \emph{Paradiso} \\
\midrule
\emph{maestro}$^{\dagger}$ & \emph{notte}$^{\dagger}$ & \emph{luce}$^{\dagger}$ \\
\emph{duca}$^{\dagger}$ & \emph{monte}$^{\dagger}$ & \emph{letizia}$^{\dagger}$ \\
\emph{disse}$^{\dagger}$ & \emph{passi}$^{\dagger}$ & \emph{affetto}$^{\dagger}$ \\
\emph{fondo} & \emph{pur}$^{\dagger}$ & \emph{santo}$^{\dagger}$ \\
\emph{loco} & \emph{cura}$^{\dagger}$ & \emph{mortali}$^{\dagger}$ \\
\emph{dissi} & \emph{lei}$^{\dagger}$ & \emph{lume}$^{\dagger}$ \\
\emph{città} & \emph{Virgilio}$^{\dagger}$ & \emph{raggio}$^{\dagger}$ \\
\emph{mena} & \emph{ivi}$^{\dagger}$ & \emph{Cristo}$^{\dagger}$ \\
\emph{gridò}$^{\dagger}$ & \emph{cammin}$^{\dagger}$ & \emph{stella}$^{\dagger}$ \\
\emph{pianto} & \emph{vera}$^{\dagger}$ & \emph{Bëatrice}$^{\dagger}$ \\
\emph{piè}$^{\dagger}$ & \emph{novo}$^{\dagger}$ & \emph{grazia}$^{\dagger}$ \\
\emph{capo} & \emph{dicea}$^{\dagger}$ & \emph{santa}$^{\dagger}$ \\
\emph{allor}$^{\dagger}$ & \emph{carro}$^{\dagger}$ & \emph{caldo} \\
\emph{denti} & \emph{buon}$^{\dagger}$ & \emph{mondo} \\
\emph{ahi} & \emph{sole}$^{\dagger}$ & \emph{segno} \\
\emph{venimmo}$^{\dagger}$ & \emph{andar}$^{\dagger}$ & \emph{mortal} \\
\emph{quelli} & \emph{mani}$^{\dagger}$ & \emph{primo} \\
\emph{spalle} & \emph{ombra}$^{\dagger}$ & \emph{riso}$^{\dagger}$ \\
\emph{lingua} & \emph{possa}$^{\dagger}$ & \emph{vero}$^{\dagger}$ \\
\emph{pena}$^{\dagger}$ & \emph{fora}$^{\dagger}$ & \emph{sarebbe} \\
\emph{cor} & \emph{pianta}$^{\dagger}$ & \emph{fede} \\
\emph{man}$^{\dagger}$ & \emph{ombre}$^{\dagger}$ & \emph{corte}$^{\dagger}$ \\
\emph{cammino} & \emph{loro} & \emph{donna}$^{\dagger}$ \\
\emph{collo} & \emph{innanzi}$^{\dagger}$ & \emph{cielo} \\
\emph{gran} & \emph{gente} & \emph{natura} \\
\emph{alcun} & \emph{caro} & \emph{sempre} \\
\emph{paura} & \emph{dir} & \emph{Beatrice}$^{\dagger}$ \\
\emph{selva} & \emph{voler}$^{\dagger}$ & \emph{gloria}$^{\dagger}$ \\
\emph{qua} & \emph{quattro} & \emph{beato} \\
\emph{prese} & \emph{color} & \emph{Pietro} \\
\emph{via}$^{\ddagger}$ & & \emph{amore}$^{\ddagger}$ \\
 & & \emph{mente}$^{\ddagger}$ \\
 & & \emph{ciel}$^{\ddagger}$ \\
\bottomrule
\end{tabular}
\end{table}

\begin{figure}[htbp]
\centering
\includegraphics[width=0.8\linewidth]{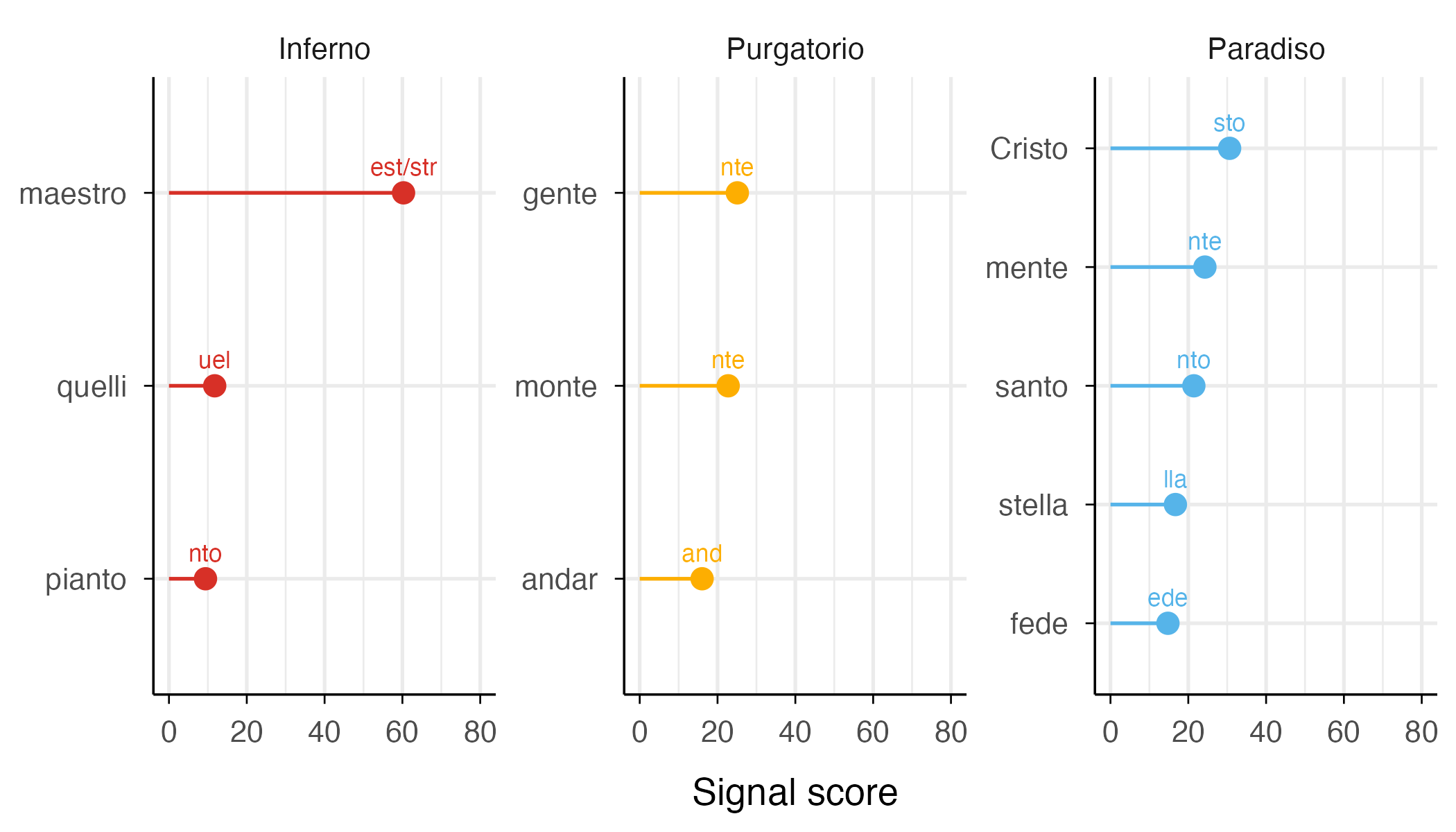}
\caption{\textbf{Lexical anchors associated with graphemic probes}. High-ranking terms identified by the classification models and containing the probes listed in Table~\ref{tab:tab-tab3}. Each point represents a lexical item positioned according to its signal score; labels indicate the associated probe.}
\label{fig:fig-fig8}
\end{figure}

To link graphemic probes to their realisations, the highest-ranking terms were examined for the presence of trigram patterns reported in Table~\ref{tab:tab-tab3}. This comparison identifies a set of anchors in which discriminative lexical signals and graphemic configurations co-occur, connecting the Markovian dependency structure to concrete textual environments. The relationship between trigram probes and their associated contexts is illustrated in Fig.~\ref{fig:fig-fig8}, where each term is positioned according to its signal score.
As noted in Table~\ref{tab:tab-tab3}, the interaction between graphemic probes and word boundaries spans a continuum. Several probes, including "ein", "ain", and "equ", display predominantly multi-word realisations (above 90\% of occurrences), whereas others, such as "ede" and "nde", exhibit intermediate behaviour, combining both intra- and inter-word contexts. By contrast, probes such as "nte", "lla", and "str" are predominantly realised within single lexical units (typically below 10\% of occurrences across word boundaries). Table~\ref{tab:tab-tab6} quantifies the distribution for this representative subset of probes. 

\begin{table}[t]
\centering
\caption{\textbf{Single-word and multi-word realisations of selected trigram probes}. Percentages correspond to the SW\% values reported in Table~\ref{tab:tab-tab3}, complemented by the corresponding multi-word proportions. The selected probes illustrate a continuum from predominantly boundary-driven to predominantly intra-lexical configurations across trigram classes.}
\vspace{0.2em}
\label{tab:tab-tab6}
\begin{tabular}{@{}l l r r@{}}
\toprule
Probe & Class & Single-word (\%) & Multi-word (\%) \\
\midrule
\emph{ein} & 1E & 0.2  & 99.8 \\
\emph{ain} & 1E & 1.0  & 99.0 \\
\emph{equ} & 2  & 1.6  & 98.4 \\
\emph{ede} & 2  & 55.3 & 44.7 \\
\emph{nde} & 1E & 75.7 & 24.3 \\
\emph{nte} & 1E & 90.5 & 9.5 \\
\emph{lla} & 1E & 92.1 & 7.9 \\
\emph{str} & 0  & 99.9 & 0.1 \\
\bottomrule
\end{tabular}
\end{table}

Representative examples further illustrate these behaviours. The probe "ein" is frequently realised across boundaries (e.g. \emph{voc\textbf{e in} sua}), while "ain" and "equ" show similarly predominant cross-word configurations (e.g. \emph{parola tu\textbf{a in}tesa}, \emph{tutt\textbf{e qu}ante insieme}). Intermediate cases, such as "ede" and "nde", occur both within words (e.g. \emph{f\textbf{ede}le}, \emph{gra\textbf{nde}}) and across boundaries (e.g. \emph{ch\textbf{e de}l bel}, \emph{a ma\textbf{n de}stra per}). Patterns such as "nte", "lla", and especially "str" are predominantly realised within single lexical units and occur in Fig.~\ref{fig:fig-fig8} within high-ranking token features such as \emph{ge\textbf{nte}}, \emph{me\textbf{nte}}, \emph{mo\textbf{nte}}, \emph{ste\textbf{lla}}, and \emph{mae\textbf{str}o}. These examples should not be interpreted as implying that the corresponding trigrams are unique to, or semantically specific to, these tokens. Rather, they illustrate the subset of cases in which independently selected graphemic probes occur within token features that are themselves discriminative in the supervised lexical analysis. Probes with intermediate behaviour can also enter this mapping, as in the case of "ede" (\emph{f\textbf{ede}}). The resulting associations therefore provide an operational correspondence between the graphemic and lexical representations, rather than evidence that individual letter sequences carry cantica-specific lexical meaning.

Taken together, these results outline a consistent relationship between local dependency structure, graphemic patterns, and lexical organisation. The Markovian representation captures systematic variation in the sequence, while the analysis of graphemic probes and lexical anchors links these statistical signals to identifiable lexical configurations.

\section{Discussion}
This study examines whether a minimal graphemic representation of Dante’s \emph{Commedia}, based on vowel–consonant encoding, can capture systematic structural variation across the poem. The results suggest a consistent relation between local graphemic dependency patterns and lexical organisation across the three cantiche, showing that even a highly reduced symbolic representation can retain meaningful structural information.

The tokeniser used in this study plays an operational rather than an interpretive role. The rule-based labels it records for apostrophe-bearing forms are not used in the present analyses; given the editorial status of apostrophes, their potential utility would instead need to be assessed comparatively across critical editions. The tokenised representation supports the lexical analyses, whereas the character-level encoding used for Markov modelling is obtained independently, by removing punctuation and non-alphabetic symbols.

A first set of results concerns the preliminary descriptive profile of the tokenised text. Mean token length shows a slight increase across the poem, while the mean number of tokens per verse decreases, with the \emph{Purgatorio} occupying an intermediate position for both measures. The two descriptors are strongly negatively associated and therefore capture closely related aspects of the distribution of lexical material within the verse. Both measures depend on the adopted tokenisation and textual representation and are therefore interpreted descriptively.

Within the Markovian framework, the analysis reveals a weak shift in local dependency structure (Fig.~\ref{fig:fig-fig4}). The two-state representation shows a clearer increase in dependence, whereas the four-state model yields a more attenuated pattern. This contrast reflects the role of model granularity: the two-state model captures a global departure from independence, while the four-state model redistributes this signal across a richer transition structure. The weaker between-cantica contrasts in the latter indicate that the observed increase in dependence is distributed across multiple local transition patterns rather than concentrated in a single component of the model. This behaviour is unlikely to be attributable to sample size effects, as the estimation of \(MD\) remains stable at the scale of individual cantos (see also Fig.~\ref{fig:fig-fig3}). More generally, the patterns detected across the different textual representations may reflect multiple compositional factors, including differences in the distribution of direct speech and the colloquial or truncated forms it tends to contain \citep{DeVentura2026}, as well as Latin or Latinising forms and proper names, whose distribution could be examined through lemmatised lexical resources such as the \emph{Vocabolario Dantesco} \citep{Manni2025}. The present analysis does not separate the respective contributions of these sources, which should therefore not be interpreted as specific mechanisms underlying the observed variation.

At the level of trigram configurations, pure persistence configurations (CCC, VVV) tend to decrease, while mixed patterns involving a single transition (CCV, VVC) increase, as do configurations in which the transition occurs at the start of the trigram (CVV, VCC). By contrast, strictly alternating configurations (CVC, VCV) show a relative decline. Within this perspective, trigram patterns can be interpreted as probes of local graphemic organisation. Some configurations display systematic monotonic behaviour across cantos, providing observable indicators of the transition patterns captured by the model (Table~\ref{tab:tab-tab3}). A key distinction emerges between probes realised within lexical units and those arising across word boundaries. Intra-word probes align more naturally with stable lexical material, whereas boundary-driven probes reflect interactions between adjacent tokens. In verse, however, such boundary-level configurations are also constrained by metrical and phonosyntactic phenomena, including sinalefe and dialefe \citep{Asperti2021}. The present graphemic representation does not distinguish among these mechanisms, and cross-word probes should therefore not be attributed to a specific stylistic process without an explicit metrical analysis.

These observations lead naturally to the question of how such configurations are reflected at the lexical level. To address this, the analysis turns to the lexical environments associated with the identified probes, using supervised classification as an interpretive tool. In this setting, classification assumes an interpretive rather than predictive role. The use of bag-of-words representations is motivated by the need for transparency and for a direct link between lexical features and the identified probes. 

In this perspective, classification performance is not an objective in itself, but provides a consistency check on the strength of the lexical signal. As shown in Table~\ref{tab:tab-tab4} and Fig.~\ref{fig:fig-fig5}, both sPLS-DA and EN-MNLR achieve a strong and highly consistent separation of the cantiche. The stable performance observed across Monte Carlo validation runs, together with the agreement between the two models, indicates that the lexical differentiation is consistently detectable under both classification frameworks. This supports the use of the classification framework as an interpretive tool.

The sPLS-DA projection (Fig.~\ref{fig:fig-fig6}) further shows that the lexical differentiation is also organised in a low-dimensional latent space. Table~\ref{tab:tab-tab5} shows how the top terms of each classifier include both widely distributed function-like items and more specific content-bearing terms. The former contribute to the overall discriminative structure through subtle but systematic shifts in frequency across cantiche, while the latter provide more direct interpretive anchors, often corresponding to recurrent textual elements within each cantica. These two components reflect complementary aspects of the lexical signal: a diffuse, distributed layer associated with frequency variation, and a more localised layer associated with lexical items.

The distribution of top-ranked lexical terms provides a more localised view of the lexical differentiation captured by the classifiers. Among the content-bearing terms, several are readily associated with recurrent textual elements within the corresponding cantica, such as \emph{denti} (teeth), \emph{spalle} (shoulders), \emph{collo} (neck) in the \emph{Inferno}; \emph{sole} (sun), \emph{carro} (chariot), \emph{passi} (steps) in the \emph{Purgatorio}; and \emph{luce} (light), \emph{letizia} (gladness), \emph{grazia} (grace) in the \emph{Paradiso}. These associations are not intended as model-derived semantic characterisations of the cantiche. Rather, they provide identifiable lexical contexts in which the graphemic patterns associated with the observed Markovian variation can be examined.

At the same time, class separation does not exhaust the structure captured by the models. Beyond discrimination among cantiche, the lexical representations show a longitudinal organisation across cantos, particularly within the \emph{Purgatorio} (Fig.~\ref{fig:fig-fig7}). At the graphemic level, the Markov representations also show canto-level variation, although the trend is clearer in the two-state model and weaker in the four-state model. These results therefore provide different, yet converging views of variation across the poem rather than evidence of a single uniform trajectory. This non-uniformity is also consistent with critical accounts in which the differentiation among the cantiche is foregrounded. \citet{Ferrante1993}, for example, characterises the \emph{Purgatorio} as transitional between the contrasting organisations of the \emph{Inferno} and \emph{Paradiso}, while showing that differences among the cantiche can involve several formal features. The present results are not intended to validate such interpretations empirically; rather, they show how some aspects of textual differentiation remain detectable, with different strength and form, even under the deliberately reduced graphemic and lexical representations considered here.

Anchored probes show a higher proportion of single-word occurrences, indicating that lexical anchoring is primarily associated with configurations realised within word boundaries. By contrast, more distributed probes are less likely to yield stable lexical correspondences. This reflects the selective nature of lexical anchoring: because classification models operate on token-level features, they preferentially capture probes that admit stable intra-word realisations, while boundary-driven patterns remain only partially represented.

This selectivity extends to the overall framework. The identification of probes introduces a structural filter, as only trigram configurations exhibiting systematic variation in line with changes in \(MD\) are retained. It is further reinforced by the choice to restrict lexical analysis to single-token representations. As a result, lexical anchoring operates under a double constraint: only configurations that are both structurally consistent with the sequence dynamics and compatible with stable intra-word realisations are retained. This restriction defines the scope of lexical anchoring: it identifies correspondences between selected graphemic probes and discriminative token features when the former admit stable intra-token realisations. Such correspondences do not imply that a probe is unique to the associated token, that the graphemic sequence itself carries semantic specificity, or that a causal relation connects the two representations.

Figure~\ref{fig:fig-fig8} further illustrates that the relation between probes and lexical items is structured. High-ranking terms often contain graphemic patterns identified as probes, with distinct distributions across cantiche. A larger number of anchors is identified for the \emph{Paradiso}. These probe--anchor correspondences should not be interpreted as evidence of a single underlying linguistic mechanism. Because lexical anchoring is restricted to single-token realisations, it preferentially captures graphemic configurations that occur within lexical units, whereas boundary-driven occurrences remain represented at the graphemic level without being forced into a lexical correspondence. The resulting pattern therefore reflects the interaction between the structural selection of graphemic probes and the token-level representation used by the classifiers.

The lexical anchors also provide interpretable points of contact with the textual organisation of the cantiche. In the \emph{Inferno}, anchors include \emph{maestro} (Virgil, the guide) and \emph{pianto} (weeping); in the \emph{Purgatorio}, \emph{monte} (mountain) and \emph{andar} (to move forward); and in the \emph{Paradiso}, \emph{Cristo} (Christ), \emph{stella} (star), \emph{santo} (holy), and \emph{fede} (faith). Alongside these, more widely distributed forms such as \emph{quelli} (those figures), \emph{gente} (people), and \emph{mente} (mind) contribute to the discriminative lexical signal. These examples are not treated as independent discoveries about the semantic organisation of the poem, but as points of correspondence between the lexical patterns identified by the classifiers and the text from which they emerge.

At the same time, the variation observed in \(MD\) remains relatively limited and is more clearly detectable in the two-state representation than in the four-state model. The weaker trend in the four-state model should not be taken as evidence that the underlying graphemic process is homogeneous or stationary.

Several limitations should be acknowledged. The analysis is based on the graphemic representation of a modern critical edition and is therefore partly edition-dependent; alternative editorial choices may affect the character sequences and token forms analysed here. Petrocchi’s critical text constitutes the fixed textual basis of the present study, and graphical features such as word division, apostrophes, and diaeresis may incorporate editorial decisions rather than directly observable authorial practices. The results should therefore be interpreted, in the first instance, as properties of the adopted textual representation. Assessing their robustness across different critical editions would provide an informative extension of the present analysis. Digital infrastructures currently under development for the manuscript and editorial tradition of the poem, such as the Naples Dante Project \citep{Ferrante2024}, may provide a basis for such comparisons, both across modern critical editions and, in perspective, across manuscript readings.

The lexical classification is restricted to token-level features and underrepresents boundary-driven configurations. The use of cantos as units provides structural coherence but limits the available sample size for inference. Finally, the V/C encoding is deliberately coarse and does not represent phonetic or metrical distinctions. These limitations point to possible extensions, including comparisons across alternative editions, richer phonological encodings, and models that explicitly account for metrical structure.

Overall, the results indicate a structured relation between local dependency patterns, graphemic organisation and lexical distribution, with related patterns remaining detectable across different representations of the text. The integration with bag-of-words classifiers further connects the low-level graphemic representation to higher-level lexical organisation.

\section{Conclusion}

This study examined the structural organisation of Dante’s \emph{Commedia} through a minimal symbolic representation based on vowel–consonant (V/C) encoding and Markov modelling. The results reveal canto-level variation in local dependency patterns, with a clearer longitudinal trend in the two-state than in the four-state representation, together with systematic variation in specific trigram configurations. Linking these configurations to their textual contexts reveals correspondences between local dependency patterns and identifiable lexical structures. Lexical anchoring provides an operational link between the graphemic and lexical representations for a subset of these patterns, while others remain associated with boundary-driven phenomena not reducible to individual lexical items.

Taken together, these findings indicate that even highly reduced representations can capture structured interactions between local dependency patterns and higher-level textual organisation. The different representations provide converging but non-uniform views of variation across the poem, rather than evidence of a single continuous trajectory. The correspondence between local dependency patterns, graphemic configurations, and lexical differentiation indicates that the structural variation detected by the reduced representation is not confined to a single analytical level. Rather, related patterns emerge across independently constructed representations of the text, providing a basis for examining their association without treating the Markovian quantities as direct measures of literary or stylistic properties.

Future work may extend this framework by incorporating richer phonological encodings, modelling cross-word dependencies more explicitly, and applying the approach to other poetic corpora. More systematic integration with machine learning approaches may further clarify the relation between local symbolic structure and higher-level textual organisation, and enable cross-linguistic comparisons within a unified modelling framework.

%
%
% Acknowledgements, contributions & roles, and references 
% will be set automatically at the end of the document
% Please note: All information in acknowledgement.tex and authors.tex 
% will be anonymized automatically for reviewing
%

\bibliographystyle{apalike}
\bibliography{references}

\end{document}